\documentclass{article}

\usepackage[preprint]{neurips_2022}

\usepackage[utf8]{inputenc} 
\usepackage[T1]{fontenc}    
\usepackage{hyperref}       
\usepackage{url}            
\usepackage{booktabs}       
\usepackage{amsfonts}       
\usepackage{graphicx} 
\usepackage{subcaption} 

\usepackage{nicefrac}       
\usepackage{microtype}      
\usepackage{xcolor}         
\title{Semi Supervised Meta Learning for Spatiotemporal Learning}

\author{%
  Faraz Waseem \\
  Stanford University\\
  \texttt{faraz333@stanford.edu} \\
  \And
  Pratyush Muthukumar \\
  Stanford University\\
  \texttt{muthukup@stanford.edu} \\
}

\begin{document}

\maketitle

\begin{abstract} 

Labeled data is hard to come by in the real world. Moreover, a majority of available data comes in the source of video and visual media. 
Recent advancements in representation learning have shown great successes in learning rich representations from a variety of inputs including text, images, and videos. 
However, these state-of-the-art architectures are data-intensive, whereas meta learning architectures possess unique capabilities of learning new tasks from diverse training tasks and corresponding labels in the few-shot regime. 
We apply semi-supervised meta learning to video data for learning spatiotemporal patterns. 
We extend work on Masked Autoencoders (MAEs) utilizing the Vision Transformer (ViT) architecture for scalable self-supervised learning in the spatiotemporal domain.  

We approached the goal of applying meta-learning to self-supervised masked autoencoders for spatiotemporal learning in three steps. 
Broadly, we seek to understand the impact of applying meta-learning to existing state-of-the-art representation learning architectures. 
Thus, we test spatiotemporal learning through: a meta-learning architecture only, a representation learning architecture only, and an architecture applying representation learning alongside a meta learning architecture. We utilize the Memory Augmented Neural Network (MANN) architecture to apply meta-learning to our framework.
Specifically, we first experiment with applying a pre-trained MAE and fine-tuning on our small-scale spatiotemporal dataset for video reconstruction tasks. 
Next, we experiment with training an MAE encoder and applying a classification head for action classification tasks. 
Finally, we experiment with applying a pre-trained MAE and fine-tune with MANN backbone for action classification tasks. 

To execute our experiments, we generate a custom small-scale video dataset of 518 human-action classes consisting of 24927 video clips and human-generated annotations sourced from the MiniKinetics-200 and TinyVIRAT datasets. We also modify the existing ViT backbone in existing MAE architectures for small-scale datasets by applying Shifted Patch Tokenization (SPT) to combats the lack of locality inductive bias available in small-scale datasets.

Our experimental results show that fine-tuning on our custom small-scale video dataset outperforms existing pre-trained MAE architectures on video reconstruction tasks. Further, we find that training an MAE encoder with a small-scale ViT backbone on our small-scale video dataset for action classification tasks converges steadily. Finally, we find that applying a pre-trained MAE and fine-tuning with an MANN backbone for action classification tasks is effective on our small-scale video dataset test tasks. 
\end{abstract}

\section{Introduction} 

Recent advancements in deep learning including the Transformer architecture have shown great success in both vision and language domains learning rich representations from a variety of inputs including text, images, and videos (\href{https://arxiv.org/abs/1706.03762}{ref: attention is all you need}). Models such as BERT have shown success in the semi-supervised regime in denoising messy data and extracting high level embeddings from partially labeled datasets (\href {https://arxiv.org/abs/1810.04805}{ref: bert}). However, real-world labeled data in the format of videos is scarce and unstructured. State-of-the-art representation learning architectures have shown great success in the vision domain in extracting high-level features from images for reconstruction or classification tasks, however, these models require massive amounts of annotated vision data. 

The field of meta learning has shown promise in learning high-level features from data in the few shot regime. Moreover, applying meta-learning to existing supervised learning architectures has been shown to allow for more data-efficient models while preserving generalizability to unseen tasks and datasets (\href{https://arxiv.org/abs/1703.03400}{ref: model-agnostic meta-learning for fast adaptation of deep networks
}). We propose applying semi-supervised meta-learning to video data for learning spatiotemporal patterns. We believe that wrapping existing state-of-the-art self-supervised representation learning architectures within a meta-learning framework will allow our architecture to both improve sample efficiency and generalize well to unseen data, particularly in the application of spatiotemporal learning on video datasets. Specifically, we perform experiments in the style of an ablation study to compare the performances of existing representation learning architectures for video data alone, existing self-supervised meta learning frameworks for video data alone, and our formulation of applying meta learning to representation learning architectures for video data classification tasks. 

In addition to considering the effectiveness of applying meta learning towards existing representation learning architectures, we perform modifications to perform experiments with the scope of this project. That is, we scale down the vision transformer (ViT) backbone within the existing representation learning architecture for training on our custom small-scale video dataset. We generate this dataset consisting of video clips describing human-object interactions as well as corresponding human-generated annotations. 

In this project, we make the following contributions:
\begin{enumerate}
    \item We collect a custom small-scale human-object video dataset built as a composite dataset from existing human-object video sources upon which we preprocess.
    \item We apply the meta-learning framework to existing self-supervised representation learning architectures and apply our model to downstream tasks including video reconstruction and action classification
    \item We perform an ablation study to understand the impact of applying meta-learning to existing self-supervised representation learning architectures on action classification accuracy and video reconstruction loss
\end{enumerate}

\section{Related Works} 

Prior work in the field of representation learning has shown successes in learning rich representations from vision and language domains. Particularly, autoencoder architectures have been proven to be effective in extracting representations from text and images. (\href{https://arxiv.org/abs/2111.06377}{ref: masked autoencoders are scalable vision learners}) proposed applying masked autoencoders (MAEs) for self-supervised learning for vision. By masking random patches of the input images and pre-training an autoencoder to reconstruct the missing pixels, they found that the architecture was able to perform well on the ImageNet dataset compared to similar self-supervised models. Moreover, their architecture was more efficient and scalable for larger models such that transfer performance in downstream tasks outperformed supervised pre-training models. They noted that a masking ratio larger than 75\% masked pixels in an image poses as a non-trivial task to current state-of-the-art vision models.

(\href{https://arxiv.org/abs/2205.09113}{ref: masked autoencoders as spatiotemporal learners}) builds off of this work by applying masked autoencoders for video data to learn spatiotemporal patterns. The masking process follows similarly from above, however random spacetime patches of videos are masked out rather than pixels during the pre-training step. Their results showed that a masked autoencoder with a masked ratio of 90\% outperforms supervised pre-training approaches by a wide margin on both benchmark datasets and real-world video data. 

Meta-learning has shown effectiveness in generalizing well to unseen data with sample-efficient architectures in the few shot regime. One such implementation of meta-learning is the Memory Augmented Neural Network (MANN) architecture proposed by (\href{https://dl.acm.org/doi/10.5555/3045390.3045585}{href: meta-learning with memory-augmented neural networks}). The authors propose a black-box meta-learning framework with a two-part architecture. Their architecture included a controller implemented as a sequence model -- they utilize an LSTM architure in their implementation -- and an external memory module with reading and writing heads implemented with a Neural Turing Machine (NTM) (\href{https://arxiv.org/abs/1410.5401}{neural turing machines}). The LSTM sequence models are used to help a model to learn quickly from data with a small number of examples.

In our review of this space, we have not found existing work applying meta-learning alone towards self-supervised spatiotemporal learning. However, prior research has been done on applying self-supervised meta learning for natural language classification tasks (\href{https://aclanthology.org/2020.emnlp-main.38/}{href: self-supervised meta-learning for few-shot natural language classification tasks}). 

Current vision models have become increasingly powerful since the widespread application of the Transformer architecture. The Vision Transformer (ViT) architecture, proposed by (\href{https://arxiv.org/abs/2103.15691}{ref: ViViT: a video vision transformer}), builds upon the self-attention mechanism proposed by (\href{https://arxiv.org/abs/1706.03762}{ref: attention is all you need}) for learning complex high dimensional representations from image datasets. This family of architectures relies on large amounts of image data, typically in the scale of hundreds of gigabytes worth of labelled images to train large architectures with hundreds of millions of parameters.

Some work has been done on scaling down these large-scale ViT architectures while preserving the learned high-level representations.(\href{https://arxiv.org/abs/2112.13492}{ref: vision transformer for small-size datasets}) proposes Shifted Patch Tokenization (SPT) and Locality Self-Attention (LSA) as methods to combat the lack of locality inductive bias available in small-scale datasets.

Existing work on applying representation learning architectures such as MAEs with ViT backbones show incredible performance in video classification and video reconstruction tasks, but are limited in real-world applications due to the data requirements of these sample inefficient architectures. Current research on small-scale ViT architectures perform well on image classification tasks, but have yet to be extended towards video data or applied in the regime of self-supervised learning. 

\section{Methods} 

We approach the goal of applying meta-learning to self-supervised masked autoencoders for spatio-temporal learning using MANNs (memory augmented neural networks), in a similar fashion proposed by (\href{https://dl.acm.org/doi/10.5555/3045390.3045585}{href: meta-learning with memory-augmented neural networks}). In our case, we utilize the masked autoencoder (MAE) approach for initial pre-training, and then fine-tune using the MANN approach, using the MAE encoder as a backbone to the sequence model. In our implementation, we utilize the ViT sequence model scaled down and trained on our small-scale video dataset. We scale down the ViT backbone within the MAE encoder and decoder in a method proposed by (\href{https://arxiv.org/abs/2112.13492}{ref: vision transformer for small-size datasets}), however in their implementation, they focus on image data.

We consider the MAE method proposed by (\href{https://arxiv.org/abs/2205.09113}{ref: masked autoencoders as spatiotemporal learners
}) as a baseline for testing the performance of a state-of-the-art classification algorithm that does not use meta learning. We then train the MANN architecture with the ViT backbone end-to-end to evaluate the performance of a solely meta-learning based approach. Finally, we test our proposed combination of MAE with MANN fine-tuning to test if the MAE architecture in combination with meta-learning approaches is more effective in learning spatiotemporal patterns.

One benefit of applying meta-learning in this domain is that if we assume videos of humans interaction with objects share some high level structure, we can combine video clips from various human-object interaction datasets, allowing us to pre-train on more data. These combinations of benchmarks will allow us to pinpoint whether applying meta-learning with MAE is effective for spatiotemporal learning as well as the individual contributions of each.

To summarize, we devised a three-stage approach to reaching our proposed goals:
\begin{enumerate}
    \item Apply pre-trained MAE and fine-tune for video reconstruction downstream task
    \item Train MANN with MAE encoder on small-scale dataset and apply classification head for action classification downstream task
    \item Apply pre-trained MAE and fine-tune with MANN backbone for action classification downstream task
\end{enumerate}

\begin{figure}[h]
    \centering
    \includegraphics[width=\textwidth]{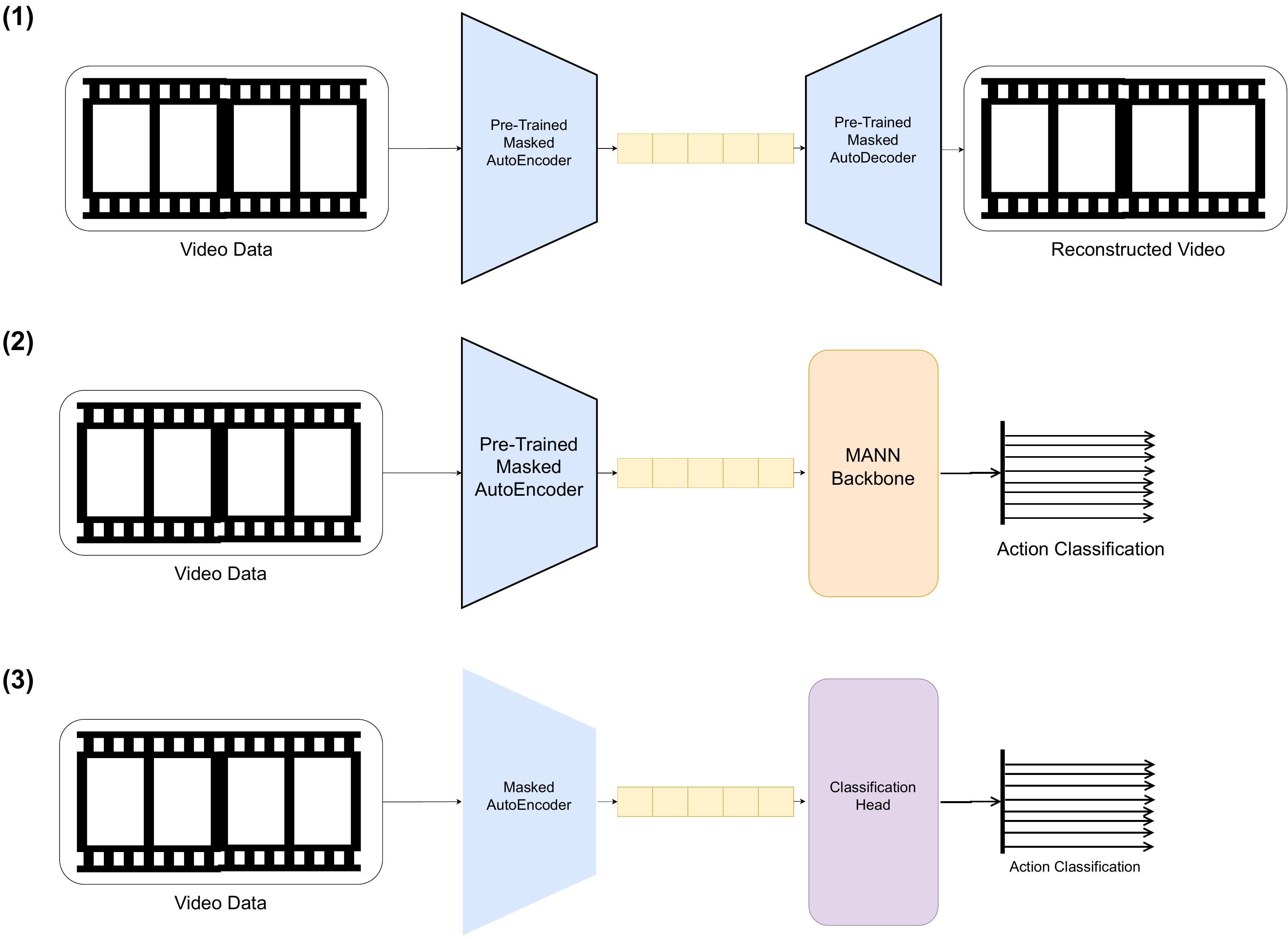}
    \caption{Visualization of Model Architectures for Each of the Three Stages in Our Approach}
    \label{threeStage}
\end{figure}

Describes a visualization of the model architectures for each of the three approaches we implement.

\section{Experiments} \label{experiments} 

For the first approach in our technical method, we fine-tune the pre-trained MAE on our small-scale dataset and evaluate against the baseline video MAE model pre-trained on Kinetics-400. We utilize a pre-trained MAE architecture sourced from the authors of the video MAE architecture trained with the ViT-Large backbone on Kinetics-400 with a masking ratio of 90\% and 1600 effective epochs (\href{https://arxiv.org/abs/2205.09113}{ref: masked autoencoders as spatiotemporal learners
}).  

\begin{table}[h]
    \centering
    \small
    \begin{tabular}{l c c}
    \hline
        \bf{Hyperparameter} \hspace{5cm} & \bf{Value}  \\
    \hline
        Input Dimension \hspace{5cm} & 64 \\
        Sampling Rate \hspace{5cm} & 4 \\
        Random Erase Mode \hspace{5cm} & False \\
        AutoAugment Policy \hspace{5cm} & rand-m7-mstd0.5-inc1 \\
        Accumulate Gradient Iterations \hspace{5cm} & False \\
        Clip Gradient Norm \hspace{5cm} & False \\
        Layer Decay \hspace{5cm} & 0.75 \\
        Base Learning Rate \hspace{5cm} & 0.0024 \\
        Drop Path Rate \hspace{5cm} & 0.1 \\
        Dropout \hspace{5cm} & 0.5 \\
        Mixup \hspace{5cm} & 0.8 \\
        Label Smoothing \hspace{5cm} & 0.1 \\
        Epochs \hspace{5cm} & 50 \\
        Warmup Epochs \hspace{5cm} & 5 \\ 
        Batch Size \hspace{5cm} & 2 \\
    \hline
    \end{tabular}
    \caption{Pre-trained MAE Fine-Tuning Parameters}
    \label{tab:hyperparameter}
\end{table}

For the second approach in our technical method, we train the MAE autoencoder with our small-scale ViT and fine-tune with a classification head on our small-scale composite dataset. We ran experiments training the full video MAE as well as training the video MAE outfitted with a classification head. Additionally, we evaluate training the video autoencoder with and without masking to analyze the difference in training loss and classification accuracy. Note that the autoencoders used for training in this set of experiments utilize our small-scale ViT backbone which implements Shifted Patch Tokenization (SPT) to preserve locality-specific representations typically lost with small-scale datasets. Further, since the original work proposing small-scale ViT architectures implemented a small-scale ViT for image classification rather than video classification, we extend upon their work by including spacetime attention to their small-scale ViT architecture in order to support 3D video data in the format of time-indexed series of 2D images.

\begin{table}[h]
    \centering
    \small
    \begin{tabular}{l c c}
    \hline
        \bf{Hyperparameter} \hspace{5cm} & \bf{Value}  \\
    \hline
        Input Dimension \hspace{5cm} & 64 \\
        Random Erase Mode \hspace{5cm} & True \\
        Random Erase Probability \hspace{5cm} & 0.25 \\
        AutoAugment \hspace{5cm} & False \\
        Accumulate Gradient Iterations \hspace{5cm} & False \\
        Clip Gradient Norm \hspace{5cm} & False \\
        Weight Decay \hspace{5cm} & 0.005 \\
        Learning Rate \hspace{5cm} & 0.001 \\
        Rate of Stochastic Depth \hspace{5cm} & 0.1 \\
        Mixup \hspace{5cm} & 0.5 \\
        Mixup Interpolation Coefficient \hspace{5cm} & 1.0 \\
        Label Smoothing \hspace{5cm} & 0.1 \\
        Epochs \hspace{5cm} & 50 \\
        Warmup Epochs \hspace{5cm} & 10 \\ 
        Batch Size \hspace{5cm} & 4 \\
    \hline
    \end{tabular}
    \caption{Hyperparameter Values for Training Video Auto Encoders from Scratch}
    \label{tab:params}
\end{table}

\subsection{Datasets} 
\begin{figure}[h]
\begin{subfigure}[h]{0.5\linewidth}
\includegraphics[width=\linewidth]{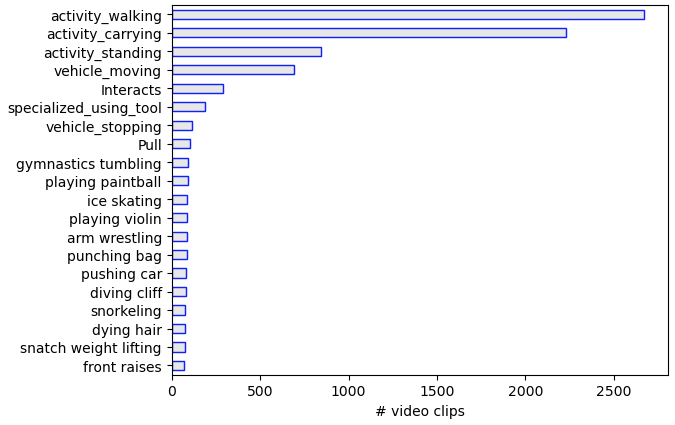}
\caption{Training Split}
\label{trainSplit}
\end{subfigure}
\begin{subfigure}[h]{0.5\linewidth}
\includegraphics[width=\linewidth]{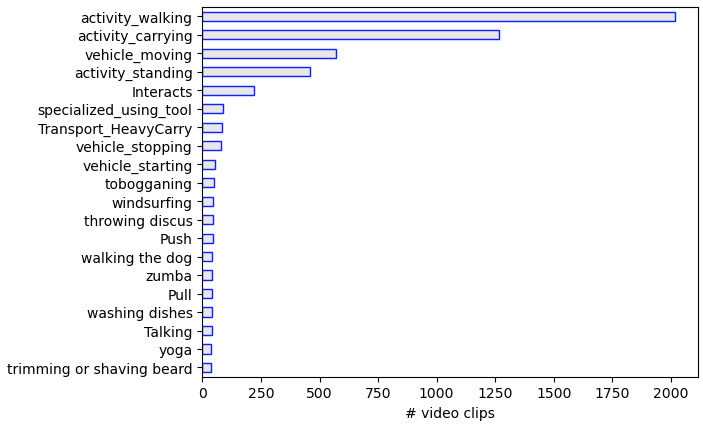}
\caption{Testing Split}
\label{testSplit}
\end{subfigure}%
\caption{Top 20 Action Classes with Most Video Clips for Training and Testing Split of Custom Small-Scale Video Dataset}
\end{figure}

For our experiments, we seek to perform spatiotemporal learning on video datasets. Initially, we started by utilizing the Kinetics-400 video dataset consisting of 400 human-action classes each with at least 400 video clips (\href{https://arxiv.org/abs/2007.07355}{TinyVIRAT: low-resolution video action recognition}). In total, the dataset consisted of 306,245 video clips each around 10 seconds in length with a resolution of 224 x 224 pixels. However, the size of this dataset is over 300 GB, and while it can be effectively used for the ViT-base backbone with 84,943,656 parameters within the MAE encoder of the existing state-of-the-art representation learning architecture for video learning, it was not a feasible dataset within the scope of our project. Instead, we developed a small-scale ViT backbone within the MAE encoder architecture which instead has 3,109,008 parameters. Correspondingly, we sought to scale down our video dataset used for training our small-scale ViT backbone.

One aspect we considered while building our dataset was since we apply the MANN meta-learning framework for self-supervised spatiotemporal learning, we can combine multiple datasets of varying action class distributions together into a composite dataset where each unique action class could be considered a new task during black-box adaptation with the MANN architecture. As a result, we were not limited to a single data source when constructing our small-scale dataset and instead, we utilized human-action video clips and annotations from a variety of input sources to generate our small-scale video dataset. In a semi-supervised dataset, labels are sparse, hence we hypothesize that a meta-learning based approach that learns quickly from a small number of examples can excel where standard fine-tuning may not be sufficient.

Our composite small-scale video dataset was sourced from the Kinetics-400, MiniKinetics-200, and TinyVIRAT datasets. MiniKinetics-200 is a subset of the Kinetics dataset consisting of the 200 human-action classes with the most training examples and TinyVIRAT is a video dataset containing real-life tiny actions in videos collected from low resolution video cameras consisting of 12829 video clips. Our small-scale video dataset contains 24,927 video clips amongst 518 human-action classes. Each video clip in our dataset consists of 100 frames at a temporal resolution of 10 FPS, meaning that each clip is around 10 seconds in length. We scale all clips in our dataset to a resolution of 64x64 pixels to perform efficient training and achieve our project goals with the computational resources available to us. All spatial and temporal resolution downscaling was performed using the OpenCV Python package.

We split our dataset into training and testing splits such that we reserve 18406 videos over 414 action classes for training and 6521 videos over 104 action classes for testing. For our implementation utilizing meta-learning for self-supervised spatiotemporal learning, each human-action class can be formulated as a distinct task, where our task training-testing split is roughly an 80-20 split. Kinetics-400, Mini-Kinetics200, and TinyVIRAT all include human-generated annotations of the video clips, which define the locations of individual video clips within the action classes.

\section{Results} 

For the first approach in our technical outline, we provide cross entropy loss results of a pre-trained MAE fine-tuned on our small-scale video dataset against a pre-trained baseline MAE architecture trained on Kinetics-400. For the sake of brevity, we provide experimental results for every 20th frame in the 100 frame video samples of our small-scale video dataset. We evaluate the pre-trained MAE baseline against our fine-tuned MAE model on the testing set of our small-scale video dataset consisting of 6521 100-frame video clips of 64x64 pixel resolution over 104 human-action classes. Table \ref{tab:crossOne} describes the averaged cross entropy loss for every 20th frame in the 100-frame video clips across the test set for our fine-tuned model compared against the pre-trained MAE baseline. The overall averaged cross entropy loss for all 100-frames across the test set in our pre-trained model was $0.1776$, whereas the pre-trained MAE baseline was $0.1781$. 

\begin{table}[h]
    \centering
    \begin{tabular}{cccccc} \toprule
    \bf{Model} &\bf{Frame 20} & \bf{Frame 40} & \bf{Frame 60} & \bf{Frame 80} & \bf{Frame 100}\\ \midrule
    Ours & 0.1785 & \textbf{0.1825} & 0.1992 & \textbf{0.1672} & \textbf{0.1457}  \\
    Video MAE & \textbf{0.1621} & 0.1948 & \textbf{0.1635} & 0.1758 & 0.1885 \\\bottomrule
\end{tabular}
\caption{Averaged Cross Entropy Loss For Every 20th Frame of 100-Frame Video Samples in Testing Split of Small-Scale Video Dataset: Ours vs Video MAE}
\label{tab:crossOne}
\end{table}

We also provide a video reconstruction visualization for a single video in the testing split of our small-scale video dataset. Since we cannot show all 100 frames of this video reconstruction, we show a visualization of every 20th video frame reconstructed by our fine-tuned model in Figure \ref{reconstruct}.

\begin{figure}[t]
    \centering
    \caption{MAE Pre-Trained on Kinetics-400 and Fine-Tuned on Small-Scale Dataset: Sample Video Reconstruction Visualization}
    \label{reconstruct}
\end{figure}

For the second approach in our technical outline, we evaluate training our modified video MAE architecture with a small-scale ViT backbone end-to-end as well as training with a classification head attached for action classification tasks. These experiments were conducted on the TinyVIRAT dataset with 26 action classes, so we can formulate the experimental setting as a 26-way multi-class classification task. The end-to-end video MAE architecture with a small-scale ViT backbone contains 3.1 million parameters, while the video MAE architecture with the classification head contains 2.7 million parameters. 

\begin{figure}[b]
\begin{subfigure}[h]{0.5\linewidth}
\includegraphics[width=\linewidth]{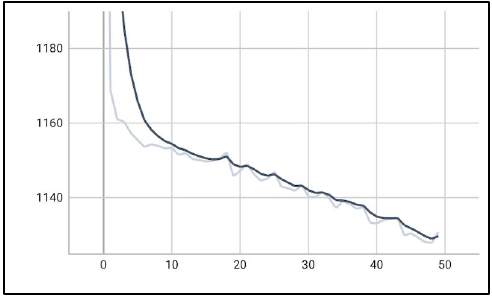}
\caption{Training Loss}
\label{fullTrain}
\end{subfigure}
\begin{subfigure}[h]{0.5\linewidth}
\includegraphics[width=\linewidth]{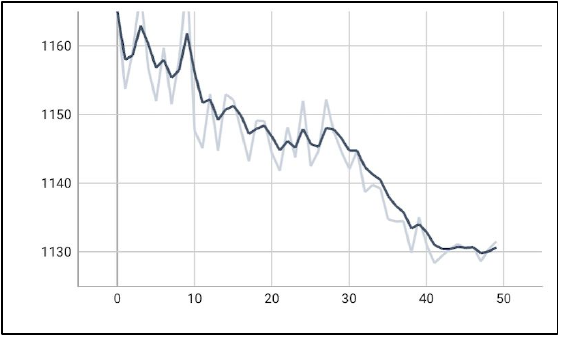}
\caption{Validation Loss}
\label{fullValid}
\end{subfigure}%
\caption{Full Video MAE Architecture with Small-Scale ViT: Training and Validation Loss Curves}
\end{figure}

The top-1 accuracy for the end-to-end video MAE architecture with a small-scale ViT backbone was $37\%$ and the top-5 accuracy was $75\%$. Figures \ref{fullTrain} and \ref{fullValid} describe the training and validation curves of this end-to-end model. Note that since we do not normalize the loss value with the number of examples in the batch, the magnitude of the loss is not necessarily indicative of the model performance. 

\begin{figure}[h]
\begin{subfigure}[h]{0.5\linewidth}
\includegraphics[width=\linewidth]{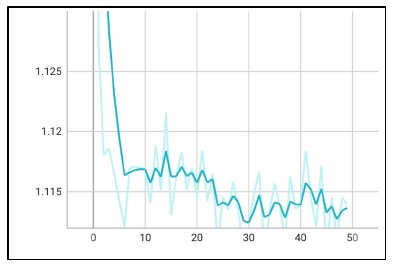}
\caption{Training Loss}
\label{maskTrain}
\end{subfigure}
\begin{subfigure}[h]{0.5\linewidth}
\includegraphics[width=\linewidth]{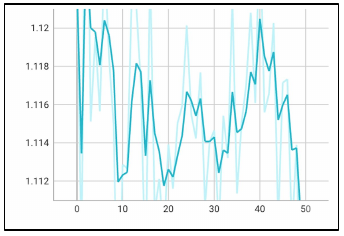}
\caption{Validation Loss}
\label{maskVal}
\end{subfigure}%
\caption{Video Masked Auto Encoder Architecture with Small-Scale ViT Training and Classification Head: Training and Validation Loss Curves}

\begin{subfigure}[h]{0.5\linewidth}
\includegraphics[width=\linewidth]{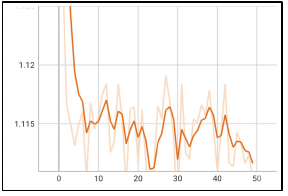}
\caption{Training Loss}
\label{noMaskTrain}
\end{subfigure}
\begin{subfigure}[h]{0.5\linewidth}
\includegraphics[width=\linewidth]{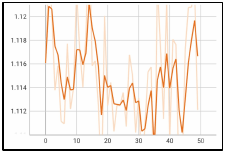}
\caption{Validation Loss}
\label{noMaskVal}
\end{subfigure}%
\caption{Video Auto Encoder Architecture with Small-Scale ViT and Classification Head: Training and Validation Loss Curves}

\begin{subfigure}[h]{0.5\linewidth}
\includegraphics[width=\linewidth]{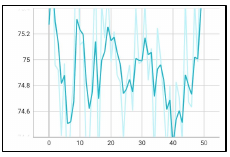}
\caption{Video Masked Auto Encoder Accuracy}
\label{accuracyMask}
\end{subfigure}
\begin{subfigure}[h]{0.5\linewidth}
\includegraphics[width=\linewidth]{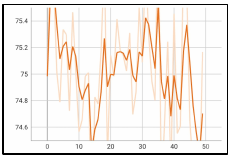}
\caption{Video Auto Encoder Accuracy}
\label{accuracyNoMask}
\end{subfigure}%
\caption{Accuracy Curves over Validation Split Through Training: Video Masked Auto Encoder with Classification Head vs Video Auto Encoder with Classification Head without Masking}
\end{figure}

Additionally, we evaluate training the video auto encoder outfitted with a classification head with and without masking for our 26-way multi-class classification task. We consider a masking ratio of $80\%$ when implementing masking. We find that the top-5 performance on the TinyVIRAT dataset is 76\% with masking and 74.5\% without masking. Figures \ref{maskTrain} and \ref{maskVal} describe the training and validation curves for the video autoencoder with a classification head with masking implemented. Figures \ref{noMaskTrain} and \ref{noMaskVal} describe the training and validation curves for the video autoencoder with a classification hea trained without masking. Figures \ref{accuracyMask} and \ref{accuracyNoMask} describe the validation split accuracy curve over training for the masked autoencoder and the autoencoder without masking, respectively.

When using a video autoencoder with shift patch tokenization, and a reduced number of parameters, in only 10 epochs of pretrainign and 10 epochs of finetuning, we get 46.8\% top1 accuracy, which is significantly higher than the previous methods we tested, indicating the importance of using shifted patch tokenization and not masking during the finetuning phase.

\section{Conclusion} 

To summarize, we apply self-supervised meta-learning for spatiotemporal learning on video data. We extend existing representation learning architectures for vision and video data and apply meta-learning through the black-box Memory Augmented Neural Network (MANN) architecture. We evaluate the effectiveness of applying MANN alongside Masked Auto Encoders (MAE) by tackling our goals for this project in a three stage approach. 

Firstly, we experiment with fine-tuning a pre-trained MAE architecture on our custom small-scale video dataset. This small-scale video dataset is built and collected by combining multiple human-action video datasets such as the TinyVIRAT, Kinetics-400, and MiniKinetics-200 datasets. Our experimental results of our fine-tuned model against a pre-trained MAE baseline shows that our model outperforms the pre-trained MAE architecture in terms of averaged cross entropy loss across all frames of the testing split videos in our small-scale dataset with a value of $0.1776$ compared to the baseline's averaged cross entropy loss of $0.1781$. However, since the difference between these two values are negligible -- our fine-tuned model outperforms the baseline by $0.3\%$ -- we note that there is not a significant enough improvement from fine-tuning a pre-trained MAE architecture on our small-scale video dataset alone. We anticipated these results and hypothesize that because the pre-trained model is very large and trained on hundreds of gigabytes worth of Kinetics-400 data, whereas we fine-tune on our small-scale dataset consisting of less than 25,000 video clips, fine-tuning this architecture directly will not have a noticeable impact on predictive power. Nevertheless, our fine-tuned model slightly outperforms the baseline pre-trained MAE architecture, however there are not enough results or significant enough a difference to suggest a trend. 

Next, we experiment with training an end-to-end video MAE architecture with a modified small-scale ViT backbone. We evaluated this architecture on the TinyVIRAT dataset and formulated the problem as a 26-way multi-class video classification problem. The top-1 accuracy score was $37\%$ and the top-5 accuracy score was $75\%$. We believe this is a significant accomplishment because the majority of existing benchmarks for the TinyVIRAT challenge utilize very large encoder architectures with hundreds of millions of parameters. However, we are able to achieve competent results on the TinyVIRAT dataset with a small-scale ViT backbone with just 3 million parameters. 

Finally, we experiment with training a video auto encoder architecture with a classification head and evaluating the effect of masking. We similarly evaluated both the masked and non-masked architectures on the TinyVIRAT 26-way multi-class video classification task and find that the top-5 performance for the masked auto encoder architecture with an 80\% masking ratio was $76\%$ and for the auto encoder without masking was $74.5\%$. Comparatively, this shows that applying masking to the architecture improves action-class classification task performance. However, with just 50 epochs used for training, we would need to continue running experiments and fine-tune the masking ratio hyperparameter to confirm this trend.

\section{Future Work} 

In the future, we want experiment with fine tuning the MANN architecture with and without a pre-trained video MAE. Another test we want to try is to replace MANN with other meta-learning implementations such as Model Agnostic Meta-Learning (MAML) proposed by (\href{https://arxiv.org/abs/1703.03400}{ref: model-agnostic meta-learning for fast adaptation of deep networks
}). We can also experiment with integrating text signals such as utilizing BERT pretrained embeddings generated on descriptions of videos in the action-class classification task setting. We have performed significant contributions to the TinyVIRAT codebase and could consider contributing to open-source implementations by providing our codebase for small-scale video MAE and meta-learning capabilities. Additionally, we have introduced a hook to export latent video frame representations, which can be used for future work by us and others. We believe we have created very useful building blocks for building more advanced vision transformers for the spatiotemporal learning domain. 



\end{document}